\documentclass[10pt,twocolumn,letterpaper]{article}

\usepackage[pagenumbers]{cvpr} % To force page numbers, e.g. for an arXiv version

\renewcommand{\paragraph}[1]{\vspace{.5em}\noindent\textbf{#1}}

\usepackage[table]{xcolor}

\definecolor{gray}{rgb}{0.9, 0.9, 0.9}
\definecolor{first}{rgb}{0.92, 0.97, 0.85}
\definecolor{second}{rgb}{1.0, 0.93, 0.7}
\def \first {\cellcolor{first}}

\def \gray {\cellcolor{gray}}

\usepackage{tikzducks}
\usepackage{bm}
\usepackage{multirow}
\usepackage{cuted}

\usepackage[most]{tcolorbox}
\tcbuselibrary{listings}

\definecolor{cvprblue}{rgb}{0.21,0.49,0.74}
\usepackage[pagebackref,breaklinks,colorlinks,allcolors=cvprblue]{hyperref}

\def\paperID{****} % *** Enter the Paper ID here
\def\confName{CVPR}
\def\confYear{2026}

\newcommand*{\ours}[0]{WHOLE\xspace}

\newcommand{\supmat}{\textcolor{magenta}{\emph{Sup.~Mat.}}\xspace}

\usepackage{longtable}
\usepackage{array}
\usepackage{fancyvrb}
\title{WHOLE: World-Grounded Hand-Object Lifted from Egocentric Videos}

\author{
Yufei Ye\textsuperscript{1} \qquad Jiaman Li\textsuperscript{2} \qquad Ryan Rong \textsuperscript{1} \qquad C. Karen Liu \textsuperscript{1,2} \\
\textsuperscript{1}Stanford University  \qquad \textsuperscript{2} Amazon FAR (Frontier AI \& Robotics) \\
{\tt \small \href{https://judyye.github.io/whole-www}{https://judyye.github.io/whole-www}}
}

\begin{document}
\maketitle

\begin{strip}\centering
\vspace{-1.1cm}
\includegraphics[width=\textwidth]{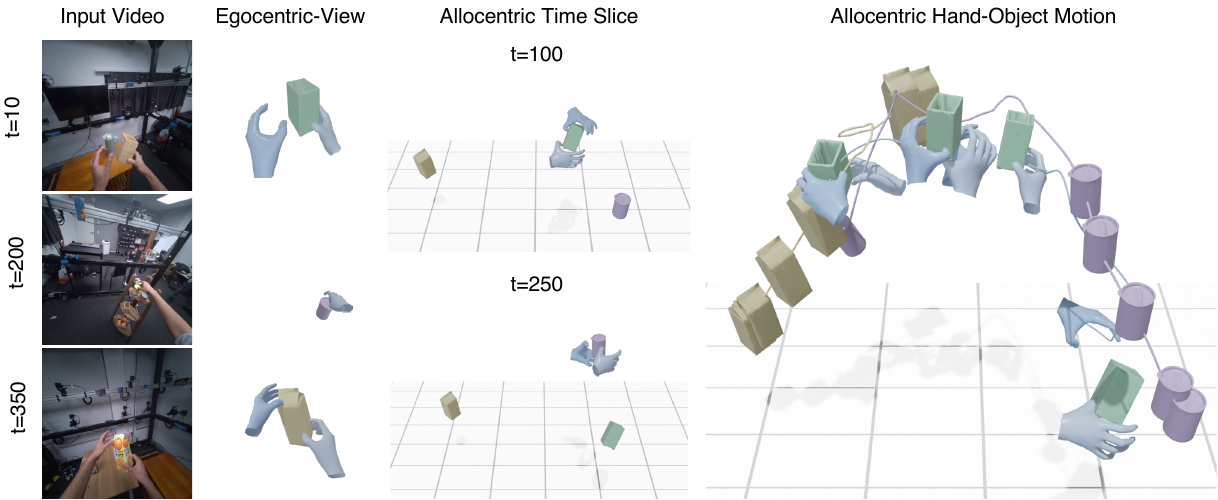}
\vspace{-0.7cm}
\captionof{figure}{Given a metric-SLAMed egocentric video of a person interacting with the scene and the corresponding object templates, \ours reconstructs the motions of both hands and the objects of interest.
The reconstruction is shown from the egocentric camera view, an allocentric view, and through the full 4D hand-object trajectories.
In the example above, the person moves a box from the table on the left to the shelf on the right, takes a can from the shelf and places it on the center table, and finally picks up an orange box.
For visual clarity, multiple objects are overlaid into a single scene, and the hands are displayed only when interacting with the box.
\label{fig:teaser}}
\end{strip}

\begin{abstract}
% goal
% challenge 
Egocentric manipulation videos are highly challenging due to severe occlusions during interactions and frequent object entries and exits from the camera view as the person moves. 
Current methods typically focus on recovering either hand or object pose in isolation, but both struggle during interactions and fail to handle out-of-sight cases. 
Moreover, their independent predictions often lead to inconsistent hand-object relations.  
We introduce \ours, a method that holistically reconstructs hand and object motion in world space from egocentric videos given object templates.
% nugget 
Our key insight is to learn a generative prior over hand-object motion to jointly reason about their interactions. 
At test time, the pretrained prior is guided to generate trajectories that conform to the video observations.  
% impact 
This joint generative reconstruction substantially outperforms approaches that process hands and objects separately followed by post-processing. 
\ours achieves state-of-the-art performance on hand motion estimation, 6D object pose estimation, and their relative interaction reconstruction.
\end{abstract}    
\section{Introduction}
\label{sec:intro}
Humans effortlessly connect what they see (egocentric) to a persistent 3D world (allocentric) - a core cognitive ability that underlies spatial reasoning and purposeful interaction.
With the increasing availability of wearable cameras, egocentric videos have become a powerful medium for capturing such first-person experiences. These recordings depict everyday activities from the wearer’s viewpoint like walking through a room, reaching for a can on the shelf, and pouring a milk bottle, as illustrated in Fig.~\ref{fig:teaser}. 
Among the many elements in these scenes, the hands and the objects they manipulate form the most direct interface through which humans act upon the world, making their 3D reconstruction a crucial step toward understanding egocentric experiences. Achieving this capacity of spatial reasoning about human interactions  enables downstream applications such as robot learning from human demonstrations~\cite{egodex,kareer2025egomimic} and immersive AR/VR environments.

Our goal is to endow computers with a comparable ability:
to reconstruct the motion of the active objects and both hands within a consistent world coordinate frame from metric-SLAMed egocentric videos that depict hand manipulation. 
However, this task is particularly challenging. Because the camera is mounted on a moving wearer, the resulting video often exhibits large egomotion even when the object itself  does not move much. Objects may leave and re-enter the field of view, and frequent occlusions between hands and objects further complicate perception and reconstruction.

% technical context, why/when they fail

Prior work has explored several closely related directions, yet often in isolation. Some methods focus exclusively on reconstructing humans~\cite{wham,slahmr,yu2025dyn} or general objects~\cite{feng2025st4rtrack,xiao2025spatialtrackerv2} in a consistent world coordinate frame, while others tackle the problem of estimating camera motion ~\cite{li2025megasam,epic_fields,patra2019ego} to align the egocentric viewpoint with the world space. However, simply combining these separate efforts is insufficient for egocentric interaction reconstruction from videos.
% , which feature rapid egomotion, severe occlusions, and objects that frequently leave and re-enter the field of view.
Another line of research addresses hand-object interaction (HOI) reconstruction ~\cite{fan2024hold,ye2023ghop,huang2022reconstructing}, typically over a few-second clips to recover detailed object geometry and contact patterns. Yet, these approaches remain confined to local reference frames, without reasoning about motion and interaction within a persistent, global 3D world.

% what we do

Our key insight is that hand and object motions are inherently interdependent, and should be modeled jointly to capture coherent hand-object interactions.
Building on this idea, we introduce \ours, which formulates reconstruction as a guided generation process based on a generative motion prior learned from hand-object interactions. Specifically, we train a diffusion-based motion prior to model the mutual dynamics between hands and objects, as well as the contact relationship between them. At test time, we guide this pretrained model using visual observations including segmentation masks and contact cues, to produce global 3D trajectories consistent with the input egocentric video.
To obtain reliable contact information, we further enhance a vision-language model (VLM) with spatially grounded visual prompts to enable robust contact localization even in cluttered scenes. Together, it allows \ours to generate coherent plausible reconstructions of long hand-object interaction sequences in global 3D space.

% what we achieve
We train and evaluate \ours on the HOT3D~\cite{banerjee2025hot3d} dataset.
We show that our learned motion prior generates diverse and plausible samples while producing reconstructions faithful to the input video.
Across hand, object, and interaction evaluation settings, \ours consistently performs well compared to existing baselines that naively combine state-of-the-art methods from the respective hand and object estimation domains.
We also find that VLM-annotated contact cues perform comparably to ground-truth labels, indicating that the designed visual prompt effectively improves the VLM’s spatial grounding. Code and model will be public upon acceptance.

\section{Related Work}
\label{sec:related}

\paragraph{Egocentric Video Understanding.}
Egocentric perception has recently gained traction due to the availability of large-scale datasets~\cite{grauman2024ego,Damen18,Ego4D2022CVPR} and wearable cameras such as GoPro and Project Aria~\cite{Lv24,Chao21,Ma24}.  
While a line of work focuses on high-level semantic understanding such as action recognition and language grounding~\cite{yang2025egovla,Ego4D2022CVPR,grauman2024ego,egodex}, recent research has expanded to spatial perception tasks, including 2D detection, segmentation, and tracking~\cite{darkhalil2025egopoints,zhao2024instance}.  
A few efforts have explored 3D understanding from egocentric inputs, such as camera localization, global human motion reconstruction, and global hand motion reconstruction~\cite{egoego,yi2025estimating,zhang2025hawor,wang2024egonav,plizzari2025spatial}. Our work follows this line of 3D egocentric understanding but differs in that we explicitly model interaction, \ie jointly reconstruct both hands, object, and their contact. This allows for reconstruction of globally coherent hand-object interactions in the world coordinate frame. 

\paragraph{Video-Based Hand Pose Estimation.}
Estimating articulated hand motion from videos has long been a fundamental problem.  
Existing methods rely on multi-view setups or additional sensing modalities such as depth or IMU data~\cite{wang2009real,garcia2018first,ruocheng_piano,h2o,arctic} to capture accurate 3D hand trajectories.  
Recent monocular approaches estimate hand poses directly from RGB videos by leveraging large-scale datasets~\cite{hamer,interhand,freihand,cocow} and transformer-based architectures~\cite{tang2021handAR,meshgraphormer,zhang2019HAMR} to regress MANO~\cite{mano} parameters.  
While these models achieve high accuracy for local hand poses, their predictions are expressed in canonical or camera-centric coordinates and thus cannot recover global trajectories in the world frame.  
To estimate global hand motion, recent works~\cite{yu2025dyn, zhang2025hawor,ye2025predicting} first predict local hand poses using a pretrained hand prior~\cite{hamer}, then infer world-space trajectories via test-time optimization or a learned neural model.  
However, these methods focus on hand motion in isolation.  
In contrast, we reconstruct object motion jointly with the hand in the world coordinate system, enabling accurate and contact-aware understanding of hand-object interactions.

\begin{figure*}
    \centering
    \vspace{-0.5em}
    \includegraphics[width=\linewidth]{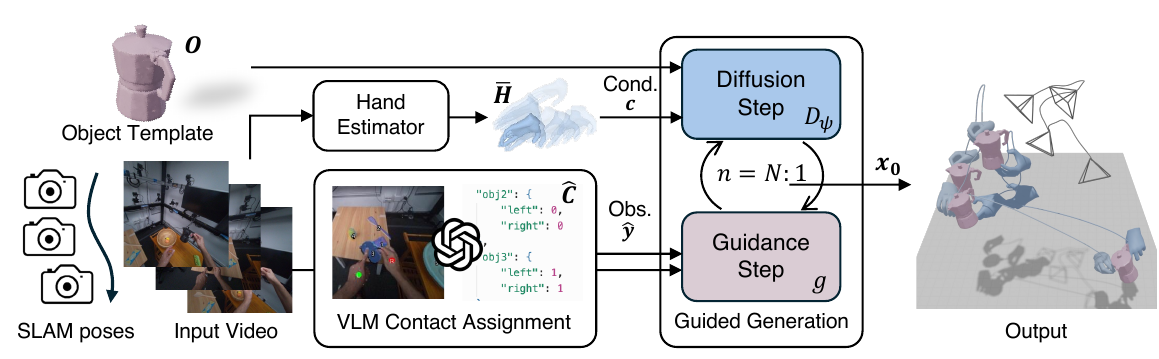}
    \caption{\textbf{Reconstruction Using the Generative Motion Prior.}  Given a metric-SLAMed egocentric videos, and the object template $\bm O$, we alternate the diffusion generation step and the guidance step to predict hand motion $\bm H$, object 6D trajectory $\bm T$, and binary contact $\bm C$ as the final output $\bm x_0$. 
    The diffusion model  $D_\psi$ is conditioned on object geometry and approximated hand $\bar {\bm H}$  from off-the-shelf hand estimator to diffuse the noisy parameters $\bm x_n$. The guidance step refines the denoised output by optimizing task-specific objectives $g$ to be consistent with the video observations $\hat {\bm y}$ like 2D masks and contact. The contact labesl $\hat {\bm C}$ is automatically labeled by prompting a VLM. 
    }
    % \kl{1. Remove "Condition" because it suggests only object template is a condition and $\bar{\bm{H}}$ is not. 2. Add $\bm{O}$ next to the object image. 3. Add $\bm{x}_0$ above the output arrow from the diffusion+guidance box. This implies that the input arrows to the diffusion box are conditions. 4. The output figure has a lot of wasted space on the right. Either crop and zoom or find another example that uses the whole space. 5. Replace "Obs." with $\bm{C}_{l,r}$ and remove $\bm{C}_{l,r}$ from the VLM box.}
    \label{fig:method}
\end{figure*}

\paragraph{Hand-Object Interaction Reconstruction.}
Beyond pose estimation, a parallel line of research reconstructs or synthesizes 3D hand-object interactions.  
Template-based methods~\cite{hasson2019learning,ho3d} estimate object pose for a given template mesh, while template-free approaches~\cite{huang2022reconstructing,wu2024reconstructing,ye2022s,ye2024g,fan2024hold} directly recover object meshes with rich shape details.  
These methods reconstruct both hand and object motion during interaction but primarily emphasize detailed geometry, typically operating in hand- or object-centered coordinates and over short temporal windows.  
Our approach instead focuses on global 4D motion reconstruction, \ie predicting how both the hand and the manipulated object move coherently through space and time, rather than recovering fine-grained surface geometry.

\paragraph{4D Reconstruction in the World Coordinate Frame.}
A related line of work aims to recover motion trajectories expressed in the world coordinate frame.  
Global human motion reconstruction methods~\cite{glamr,trace,wham,pace,ye2023decoupling} estimate 3D body trajectories in world space by decoupling human and camera motion or leveraging learned motion priors.  
In the broader 4D tracking domain, approaches such as STaRTrack~\cite{feng2025st4rtrack} and SpatialTracker~\cite{xiao2025spatialtrackerv2} perform 3D-aware point tracking to capture scene-level dynamics and camera motion, without explicitly modeling human or object structure.  
FoundationPose~\cite{foundationposewen2024} instead tackles 6D object pose estimation given known object geometry, achieving robust per-frame alignment. Our approach also assumes an input object template but further estimates object poses and hand motion jointly from videos, capturing their coupled trajectories over time.

\section{Method}
Given a metric-SLAMed egocentric video of object manipulation and a 3D object template, as shown in Figure~\ref{fig:method}, \ours holistically reconstructs both hand articulation and object 6D trajectories in world space.
We first learn a generative motion prior of hand-object interactions in a gravity-aware local frame. At test time, we guide this pretrained prior by visual observations, object masks and VLM-derived binary contact cues, to generate globally consistent motions faithful to the input video.

\subsection{Generative Hand-Object Motion Prior}
\label{sec:prior}

The generative prior is a diffusion model conditioned on a roughly estimated hand trajectory $\bar{\bm H}^{t=1:T}$ and an object template $\bm O$, \ie $\bm c \equiv (\bar{\bm{H}}, \bm O)$, generating refined hand motions $\bm H^{t=1:T}$, object trajectories as SE(3) transforms $\bm T^{t=1:T}$, and contact labels $\bm C^{t=1:T}$, two binary indicators denoting contact with the left and right hands. It operates on a fix-length time window ($T=120$) and models how the object moves and interacts with the hands in a local gravity-aware coordinate frame, \ie $p(\bm H, \bm T, \bm C \mid \bm O, \bar{\bm H})$.
The approximated hand poses $\bar{\bm H}$ are obtained from an off-the-shelf hand estimator~\cite{zhang2025hawor} at test time, while during training, we also synthesize noisy hand tracks to avoid overfitting to a particular estimator.

 \paragraph{Interaction Motion Representation.}
We represent each hand using MANO parameters~\cite{mano}, including global orientation $\bm{\Gamma}$, translation $\bm{\Lambda}$, articulation and shape $\bm{\Theta}, \bm{\beta}$, and joint positions and velocities $\bm{J}, \dot{\bm{J}}$, \ie $\bm{H} \equiv (\bm{\Gamma}, \bm{\Lambda}, \bm{\Theta}, \bm{J}, \dot{\bm{J}})$.
We concatenate the left and right hand features and use the same representation for the approximated hand trajectory $\bar{\bm H}$.
We represent object poses in SE(3) using the 9D formulation~\cite{zhou2019continuity}, and encode the (conditioning) object geometry with a BPS descriptor~\cite{prokudin2019efficient} in its canonical frame, $\bm O \equiv BPS(O)$.

To better capture the spatial relationship between the hands and the object and to encourage the diffusion model to reason about fine-grained contact, we also compute a per-time “Ambient Sensor” feature~\cite{taheri2024grip,zhang2025bimart} between them.
For each hand joint, we measure its displacement to the nearest point on the object surface transformed by the diffused pose $\bm{T}_i[O]$, which is equivalent to a BPS of the posed object using hand joints as the vector basis \ie $BPS_{\bm J}(\bm{T}_i[O])$.

\paragraph{Gravity-Aware Local Coordinate Frame.}  
Hands and objects are expressed in the gravity-aligned camera coordinate frame at the beginning of each sequence~\cite{yi2025egoallo,shen2024world}. Anchoring every sequence to the physical up direction and a consistent facing orientation allows our model to focus on relative hand-object motion rather than arbitrary global rotations.
Specifically, we rotate the camera coordinates so that the z-axis aligns with the gravity vector. These canonicalized segments can later be transformed back to world coordinates and seamlessly stitched into long, continuous sequences during guided reconstruction.

\paragraph{Approximating Inaccurate Hand Estimation.}  
Our diffusion model refines approximated hand estimates $\bar{\bm H}$  at test time by reasoning about hand-object interactions.
To enhance robustness and avoid overfitting to a specific off-the-shelf estimator, we synthesize imperfect \textit{conditioning} hand tracks during training by perturbing ground-truth MANO parameters.
We inject both trajectory-level noise $\bm{\varsigma}^g$ and per-frame noise $\bm{\varsigma}^t$ into the MANO parameters, then apply forward kinematics to produce perturbed joint positions that mimic real tracking noise, $\bar{\bm H}_{\varsigma^g, \varsigma^t}$.
We further simulate occlusion and truncation by randomly dropping frames, and train jointly on both synthesized and real estimated hand conditions.

\paragraph{Training Objective. } 
The diffusion model learns to iteratively denoise from Gaussian noise $\bm z$  to generate clean interaction trajectories $\bm x_{n=0}$.  
We adopt the conditional DDPM loss~\cite{ho2020denoising}, training the denoiser $D_\phi$ to recover clean trajectories from the corrupted ones $\bm x_{n}$:
\begin{equation}
    \mathcal{L}_{\text{DDPM}} = 
    \mathbb{E}_{t,\,\boldsymbol{\epsilon}}\!\left[w_n \,\bigl\| 
    \tilde{\bm x}_0 - D_\phi(\bm{x}_n \mid n, \bar{\bm{H}}, \bm{O})\bigr\|_2^2\right],
\end{equation}
where $\bm{x}_n$ is the diffusion parameters corrupted by Gaussian noise at step $n$ and $\tilde{\bm x}_0$ is the ground truth. $w_n$ is the variance schedule weight. The denoiser is implemented as a transformer decoder.

In addition to the DDPM loss, we introduce auxiliary objectives to enhance realism. 
(1) \textit{Interaction Loss} ($\mathcal{L}_{\text{inter}}$) encourages realistic hand-object contact by penalizing distances and ensuring near-rigid transport of contact points ~\cite{li2023object}.  
(2) \textit{Consistency Loss} ($\mathcal{L}_{\text{const}}$) promotes agreement between predicted hand features and their MANO forward kinematics.
(3) \textit{Temporal Smoothness}  ($\mathcal{L}_{\text{smooth}}$) further penalizes large accelerations.  
We first warm up the model for 10k steps using only $\mathcal{L}_{\text{DDPM}}$ before adding auxiliary terms.  Please refer to appendix for detailed implementation. 

\begin{figure}
    \centering
    \includegraphics[width=\linewidth]{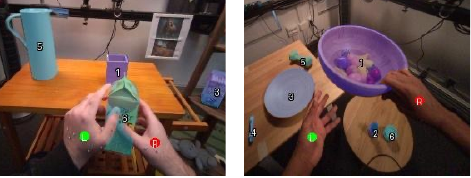}
    \caption{\textbf{Visual Prompt:} We show two examples of the visual prompts provided to the VLM for contact detection.}
    % For each frame we build a visual prompt by overlaying instance masks and index numbers on all candidate objects and placing ``L''/``R'' markers on the hands. A VLM is asked to output, for each hand, the index of the object it is in contact with (or \texttt{none} if no contact exists). \textbf{Left:} both hands grasp the same object (ID~6). \textbf{Right:} only the right hand contacts the bowl (ID~1) while the left hand is not touching any object.}
    
    \label{fig:vlm}
\end{figure}

\subsection{Reconstruction as Guided Generation}
Sampling from the learned generative prior yields diverse and realistic hand-object motions. 
We leverage this motion prior for reconstruction from monocular videos by guiding the generation through classifier-guidance~\cite{dhariwal2021diffusion}. 
Compared with another common guiding paradigm, score distillation sampling (SDS)~\cite{poole2022dreamfusion}, classifier-guidance is faster, requiring only a single forward generation pass instead of thousands of optimization steps, and is less prone to model collapse.
We find that two observations from the input video are crucial for guiding reconstruction: 2D masks that segments object and hand, and contact information indicating whether each hand is in contact with the object at the current time step.
We employ a vision-language model (VLM)~\cite{openai2023gpt} to obtain contact information.

\begin{figure*}
    \centering
    \vspace{-1em}
    \includegraphics[width=\linewidth]{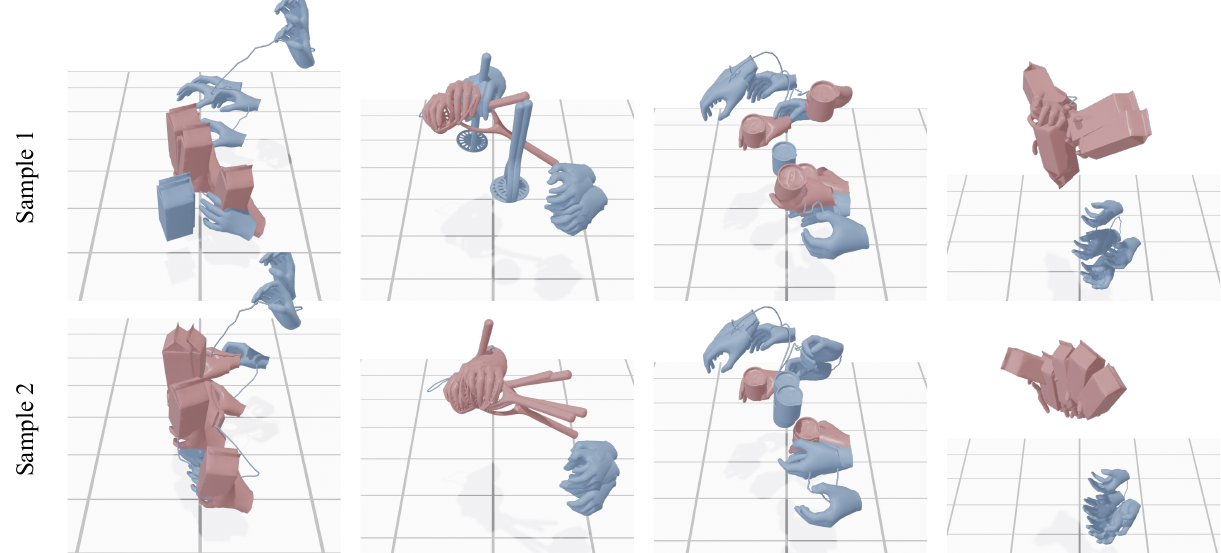}
    \vspace{-1em}
    \caption{\textbf{HOI Generation Samples:} We show two samples of interaction generated from our diffusion model with the same conditions. Objects are colored in red when contact is predicted and blue otherwise. We show 6 key frames among blended 150-frame generation. 
    % \kl{How about we give each hand and each object its own color? Try to make one hand light gray and the other darker gray. And give the objects brighter colors. I wonder how it looks. I'd also experiment with heavier shadow to see if it looks better.}
    } 
    \vspace{-1em}
    \label{fig:gen}
\end{figure*}

\paragraph{Test-Time Guidance}
Classifier guidance steers a diffusion model by modifying its score, the gradient of the log-probability $\nabla_{\bm{x}_n} \log p(\bm{x}_n)$, to incorporate task-specific objectives $g(\hat {\bm y}, \bm x)$, $\hat {\bm y}$ are observations from the video.  
During sampling, the diffusion model predicts a score via the network $D_\psi$, which is then adjusted using the gradient of the task-specific objective  $g$:
\[
\tilde{\nabla}_{\bm{x}_n} \log p(\bm{x}_n \mid \hat {\bm{y}})
= \nabla_{\bm{x}_n} \log p(\bm{x}_n)
- w \nabla_{\bm{x}_n} g(\hat {\bm{y}}, \bm{x}_n),
\]
This allows the generation process to remain plausible with respect to the learned prior while being steered toward samples faithful to the additional observations in the input videos. 

For the reconstruction task, we use three categories of objectives:
(1) Reprojection term $g_{\text{reproj}}$, which aligns the generated sample with 2D observations, including contact binaries, 2D hand joints, and object masks. We apply a one-way Chamfer loss between the reprojected object and the segmented mask to handle occlusion and truncation;
(2) Interaction term $g_{\text{inter}}$, which enforces realistic hand–object dynamics by minimizing distances, encouraging rigid transport under contact, and penalizing motion when contacts are absent in consecutive frames, following a similar formulation as $\mathcal{L}_{\text{inter}}$ during model training;
(3) Temporal smoothness term $g_{\text{temp}}$, which regularizes temporal consistency and ensures smooth trajectories.

\paragraph{VLM Contact Assignment. }
To detect hand-object contact, we use a Vision-Language Model (VLM)~\cite{team2025gemini} enhanced by spatial prompting and in-context learning. We segment and index hands and candidate objects, overlaying masks on the image to improve localization in cluttered scenes (Fig.~\ref{fig:vlm}). The VLM then identifies contacts via a JSON output, constrained by validation rules—such as a ``one-out-of-$k$" contact limit—to distinguish true touch from proximity. As we find a tendency for  false positives in VLM, we provide five annotated examples for calibration. These refinements increased the contact detection F1 score from 57\% to 81\%. For efficiency, we subsample videos to 3 fps and compute the reprojection term only at these frames.

\paragraph{Blending Long Videos.}
Our diffusion model generates motion clips within a fixed temporal window of 120 frames. 
To reconstruct sequences longer than this window ($L > 120$), we divide the sequence into overlapping sliding windows. 
During each diffusion step, we denoise all windows in parallel and blend overlapping regions~\cite{bar2023multidiffusion} and per-frame shape parameters into shared ones.
This ensures smooth temporal transitions and consistency across windows. 
The blended full-length sequence is then refined under the guidance of the cost term $g$, after which each window’s posterior $\bm{x}_n$ is updated and the diffusion process continues.

\section{Experiment}

We first train our diffusion model on HOT3D-CLIP~\cite{banerjee2025hot3d} and visualize its generated motions in Section~\ref{fig:gen}. 
We then evaluate reconstructed motions both quantitatively and qualitatively on held-out sequences, comparing \ours with state-of-the-art baselines for hands, objects, and combined baselines formed by pairing the best-performing hand and object methods. In addition, we create specific subsets of the test set, including frames where objects are in contact with the hands, truncated, or out of view,
We report reconstructed hand motion, object motion, and interaction motions both qualitatively and quantitatively (Section \ref{sec:recon}). Lastly, we analyze the effect of important components of our model (Section \ref{sec:ablation}).

\paragraph{Training Data and Setups.}
HOT3D is an egocentric dataset captured via Aria Glasses~\cite{engel2023project}, featuring real-world hand-object interactions. We utilize HOT3D-CLIP, a curated subset with verified annotations for hand-object poses, templates, and camera trajectories from shipped gravity-aware metric SLAM system.
Each sequence consists of 150 frames (3 seconds). We  train our diffusion model $D_\psi$ on 2,443 sequences. For evaluation, we hold out 50 dynamic object trajectories (displacement $> 5$ cm) from unseen sequences. While training contact labels are defined by proximity ($< 5$ mm), reconstruction labels are provided by prompting GPT-5~\cite{openai2023gpt}.

\subsection{Visualizing Hand-Object Motion Generation}
\label{sec:gen}
We visualize conditional blended generation from our diffusion model in Figure~\ref{fig:gen}.
The conditioning hand motion $\bar{\bm H}$ is taken from the test split, and we display six key frames from the full 150-frame sequence.
Meshes are shown in red when hand-object contact is predicted.
The two rows illustrate different generated samples conditioned on the same approximated hand motion.

The generated hand-object motions are realistic and physically coherent.
Within each frame, the spatial relationships between hands and objects are well captured, and the object dynamics appear plausible-objects stay relatively still when no contact is predicted and move naturally with the hands when contact occurs.
The generated hand motion are similar as they just refine the approximated hand motion. 
Yet, the predicted contact timings vary, revealing different grasp and release moments.
As a result, the reconstructed object trajectories also show diversity. For instance, in the second column, the mixer is picked up later in the first row.
The hand-object relative poses are likewise diverse, such as the bottle being grasped in different ways in the last column.
Please refer to the videos in the \supmat for the full generated motion dynamics.

\begin{table}[t]
\footnotesize
\begin{center}

\label{tab:arctic_exo}
\setlength{\tabcolsep}{2pt}
\resizebox{\linewidth}{!}{
\begin{tabular}{l c c c c c c } 
\toprule
  & WA-MPJPE $\downarrow$ & W-MPJPE $\downarrow$ & ACC-NORM $\downarrow$ & PA-MPJPE $\downarrow$ \\
\midrule
HaMeR \cite{hamer} & 16.93 & 28.35 & 32.31 & 12.76 \\
% HaMeR & \todo{run} \\
% HaPTIC & \todo{run} \\
HaWoR \cite{zhang2025hawor} & 3.76 & 11.26 & 4.15 & 8.99 \\
FP+HaWoR-simple & 3.34 & \first \textbf{9.16} & 0.95 & 8.99 \\
FP+HaWoR-contact & 5.85 & 12.19 & 1.26 & 8.99 \\
\ours (Ours) & \first \textbf{3.26} & 10.41 & \first \textbf{0.58} & \first \textbf{6.67} \\
\bottomrule
\end{tabular}
}
\vspace{-1em}
\caption{\textbf{Hand Motion.} We evaluate our method with state-of-the-art models in hand motion reconstruction. WA/W-MPJPE are in $cm$, PA-MPJPE is in $mm$. }
\vspace{-1em}
\label{tab:hand}
\vspace{-1.5em}
\end{center}
\end{table}

\begin{table*}[t]
\footnotesize
\begin{center}

\setlength{\tabcolsep}{4pt}
\resizebox{0.85\linewidth}{!}{
\begin{tabular}{l c c c c c c ccc} 
\toprule
Split & \multicolumn{3}{c}{All} & \multicolumn{2}{c}{Contact} & \multicolumn{2}{c}{Truncated} & \multicolumn{2}{c}{Out-of-view} \\
\cmidrule(r){2-4} \cmidrule(r){5-6} \cmidrule(r){7-8} \cmidrule(r){9-10} 
  & ADD $\uparrow$ & ADD-S $\uparrow$ & ACC $\downarrow$  & ADD $\uparrow$ & ADD-S $\uparrow$ & ADD $\uparrow$ & ADD-S $\uparrow$ & ADD $\uparrow$& ADD-S $\uparrow$ \\
\midrule
FoundationPose(FP)~\cite{foundationposewen2024} & 36.1 & 51.9 & 0.579 & 46.6 & 66.6 & 18.6 & 33.8 & 4.8 & 19.6 \\
FP+HaWoR-simple & 30.4 & 45.0 & 0.896 & 38.6 & 57.5 & 19.4 & 28.9 & 7.8 & 13.1 \\
FP+HaWoR-contact & 31.5 & 46.9 & 0.572 & 42.3 & 62.4 & 22.7 & 35.8 & 8.9 & 17.7 \\
Ours & \first \textbf{51.1} & \first \textbf{69.9} & \first \textbf{0.11} & \first \textbf{55.8} & \first \textbf{77.3} & \first \textbf{42.5} & \first \textbf{60.2} & \first \textbf{28.8} & \first \textbf{43.8} \\
\bottomrule
\end{tabular}
}
\vspace{-1em}
\caption{\textbf{Object Motion.} We evaluate our method with state-of-the-art baselines in object motion reconstruction.  We report AUC of ADD, ADD-S on the whole test split  (All) and 3 subsets (Contact, Truncated, and Out-of-View). }
\vspace{-1em}
\label{tab:obj}
\vspace{-1.5em}
\end{center}
\end{table*}

\begin{table*}[t]
\footnotesize
\begin{center}

\setlength{\tabcolsep}{2pt}
\resizebox{0.8\linewidth}{!}{
\begin{tabular}{l c c c c c c ccc} 
\toprule
Split & \multicolumn{3}{c}{All} & \multicolumn{2}{c}{Contact} & \multicolumn{2}{c}{Truncated} & \multicolumn{2}{c}{Out-of-view} \\
\cmidrule(r){2-4} \cmidrule(r){5-6} \cmidrule(r){7-8} \cmidrule(r){9-10} 
  & ADD $\uparrow$ & ADD-S $\uparrow$ & ACC $\downarrow$  & ADD $\uparrow$ & ADD-S $\uparrow$ & ADD $\uparrow$ & ADD-S $\uparrow$ & ADD $\uparrow$& ADD-S $\uparrow$ \\
\midrule
FP+HaWoR-simple & 29.0 & 44.0 & 0.896 & 36.8 & 55.7 & 18.6 & 28.8 & 8.1 & 16.0 \\
FP+HaWoR-contact & 36.8 & 53.8 & 0.572 & 49.1 & 71.1 & 27.3 & 43.0 & 9.8 & 21.3 \\
% Ours & \first \textbf{53.5} & \first \textbf{72.3} & \first \textbf{0.11} & \first \textbf{58.0} & \first \textbf{79.4} & \first \textbf{44.3} & \first \textbf{62.7} & \first \textbf{28.0} & \first \textbf{43.2} \\

\bottomrule
\end{tabular}
}
\vspace{-1em}
\caption{\textbf{Interactions Error.} We evaluate our method with state-of-the-art baselines in object motion reconstruction.  We report AUC of ADD, ADD-S on the whole test split  (All) and 3 challenging subsets after aligning object motion with predicted hand motion.}
\vspace{-1em}
\label{tab:hoi}
% \vspace{-1.5em}
\end{center}
\end{table*}

\begin{figure*}
    \centering
    % \vspace{-2em}
    \includegraphics[width=\linewidth]{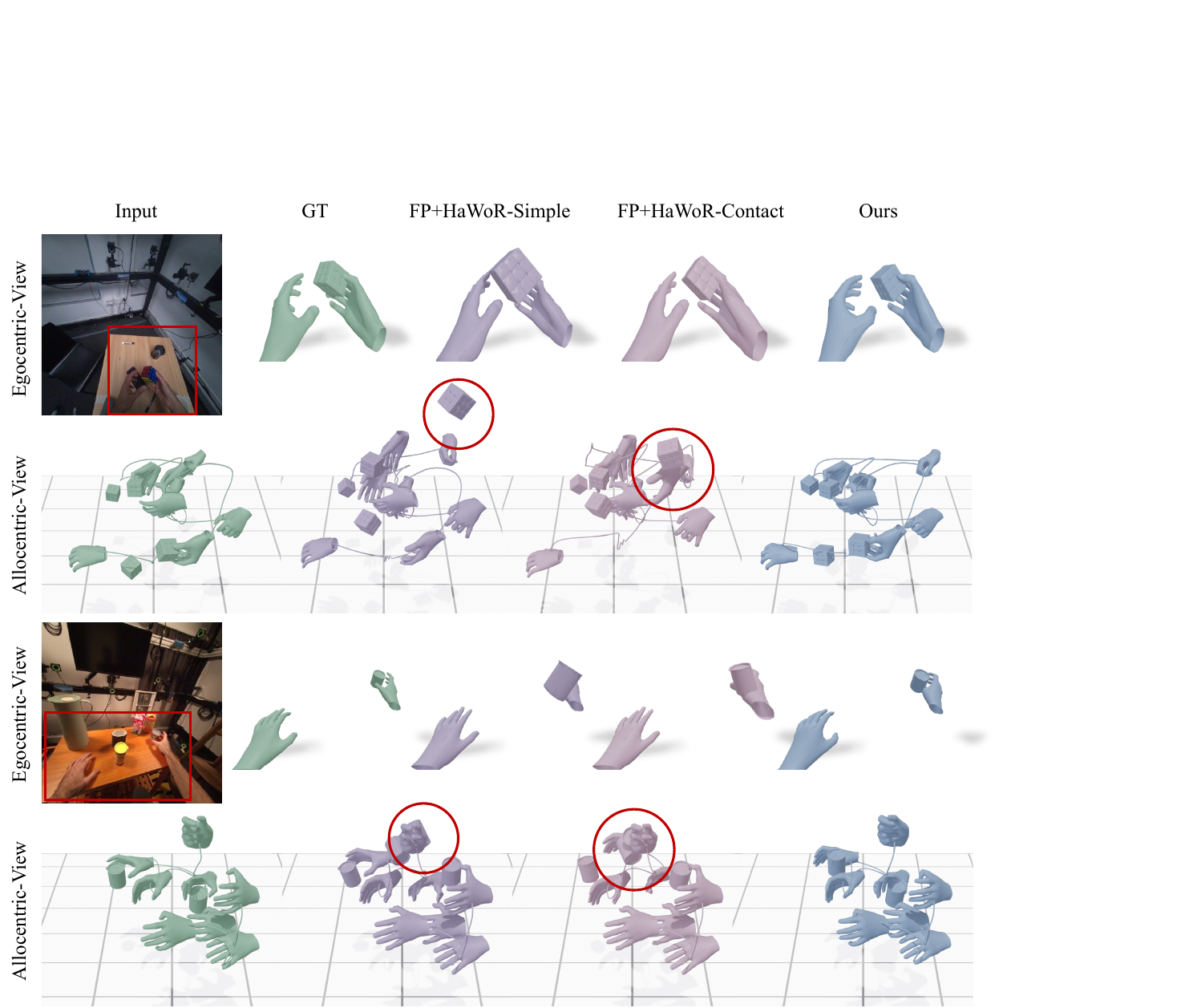}
    \vspace{-2em}
    \caption{\textbf{HOI Visualization.}  We show hand-object reconstructions from GT (green), FP+HaWor-Simple (purple), FP+HaWor-Contact (pink), and \ours (blue). Red circle highlights floating objects. We encourage readers to see videos in \supmat. }
    \label{fig:hoi}
    \vspace{-1em}
\end{figure*}

\subsection{Guided Reconstruction}
\label{sec:recon}

\paragraph{Metrics.}
\textit{Hands} are evaluated using standard metrics from whole-body/hand pose estimation~\cite{wham,slahmr,zhang2025hawor,yu2025dynhamr}.
W-MPJPE aligns all hand joints globally, WA-MPJPE aligns using the first frame, and PA-MPJPE performs per-frame Procrustes alignment.
ACC-NORM measures joint acceleration error, reflecting temporal smoothness.
\textit{Object poses } are evaluated using common metrics from 6D pose estimation: AUC of ADD and ADD-S~\cite{xiang2017posecnn,hodavn2020bop,foundationposewen2024}.
ADD measures the average distance between corresponding model points transformed by the predicted and ground-truth poses, while ADD-S extends this metric to handle symmetric objects.
To evaluate \textit{hand-object interaction}, \ie, their relative spatial alignment, we report the AUC of ADD and ADD-S for object poses \textit{after} globally aligning them using the predicted hand trajectories.

\paragraph{Baselines. } 
We compare \ours with state-of-the-art methods from their respective domains: HaWoR~\cite{zhang2025hawor} for world-grounded hand motion estimation and Foundation Pose (FP)~\cite{foundationposewen2024} for 6D object pose estimation.
Since FP requires RGB-D input, we predict depth maps from RGB videos using Metric3D~\cite{yin2023metric3d} for metric depth estimation. (We also experiment with more recent depth estimation methods~\cite{piccinelli2025unidepthv2,yang2024depth} but observe surprisingly degraded performance.)

To further evaluate joint reconstruction, we introduce combined baselines that integrate the best components from both domains, followed by test-time optimization: FP+HaWoR-simple/contact.
These baselines use the same optimization objectives $g_{\text{reproj}}, g_{\text{inter}}, g_{\text{temp}}$ as our test-time guidance (Sec.~\ref{sec:recon}).
FP+HaWoR-simple optimizes hand and object trajectories with respect to observed evidence but excludes the interaction term $g_{\text{inter}}$, while FP+HaWoR-contact includes it to enforce hand-object consistency.

\subsubsection{Comparing Hand Motion}
HaMeR is an image-based hand prediction system thus its 4D trajectories in world coordinate are not aligned well. 
We find that HaWoR performs reasonably well in terms of hand motion quality.
Note that the difference from the results reported in their paper comes from our evaluation on longer clips (150 frames instead of 100). 
Joint optimization without considering object trajectories with object motion in FP+HaWoR-simple yields smoother trajectories (lower ACC-NORM) and better global alignment. 
Adding the interaction loss (FP+HaWoR-contact) improves object performance (Table~\ref{tab:obj}, \ref{tab:hoi}) but reduces hand accuracy, revealing a trade-off between precise hand poses for stronger hand-object consistency.

In contrast, \ours achieves the best overall performance (second best on one metric), excelling in global alignment, temporal smoothness, and local pose accuracy.
Importantly, our method shows a clear improvement in local hand pose quality over HaWoR, highlighting the advantage of jointly reconstructing hands and objects within a unified generative framework.

\subsubsection{Object Motion.}
FoundationPose (FP) performs moderately well when objects are in contact but struggles under truncation or out-of-view conditions.
Note that for out-of-view frames, we interpolate 6D poses between the nearest visible frames in world space.
Optimizing baseline only with reprojection and temporal smoothness  (FP+HaWoR-simple)  helps propagate poses to out-of-view frames but reduces accuracy on the contact subsets. 
It is likely because severe hand-object occlusion-reprojection loss on occluded masks can be misleading and ultimately degrades overall performance.
Adding the interaction term (FP+HaWoR-contact) improves contact performance by leveraging hand cues but still falls short of the off-the-shelf FoundationPose.

In contrast, \ours achieves consistently strong results across all subsets and metrics.
It handles truncation and occlusion more robustly, producing smoother and more coherent object trajectories.
By jointly reasoning about hand and object motion through a learned generative prior, it infers plausible movement even when parts of the object are missing or out of view.
This demonstrates that our unified approach is more reliable than simply combining off-the-shelf methods with post-optimization.

\subsubsection{Interaction Quality. }
While we report hand and object motion separately, we also evaluate their \textit{relative motion} by measuring object error after alignment with the predicted hand trajectory.
The combined baseline (FP+HaWoR-contact) shows clear improvement over its variant without interaction optimization, highlighting the importance of modeling hand-object coupling.
Finally, \ours surpasses both baselines by a large margin, demonstrating its strong capability to capture coherent and physically consistent hand-object interactions. 

We present visual comparisons in Figure~\ref{fig:hoi}, which reflect trends consistent with the quantitative results.
In the top row of each example, we visualize the reconstructed hands and objects from the camera view, while the bottom row shows hand-object motion from an allocentric perspective across four keyframes.
As shown, baseline methods often produce floating objects (highlighted in red) and unrealistic hand-object relations, particularly in the FP+HaWoR-simple variant.
In contrast, \ours yields temporally smooth and spatially coherent reconstructions, capturing more natural and physically plausible hand-object interactions.

\subsection{Ablation Study}
\label{sec:ablation}

\paragraph{How good is VLM-annotated contact label? } 
In Table~\ref{tab:ablation}, we replace the VLM-annotated contact labels at 3 fps with ground-truth contact annotations at 30 fps.
While there remains some room for improvement, the VLM-based results is close to the ceiling performance across hand motion, object motion, and their relative spatial alignment.

\paragraph{Is intertwined generation and optimization necessary?}
\ours alternates diffusion steps with task-specific guidance throughout the generation process.
To examine the importance of this coupling, we compare it with a ``Gen+Opt" variant that first performs unguided generation, then applies post-hoc optimization using the same objective terms.
Results in Table~\ref{tab:ablation} Line 2 show that incorporating guidance \textit{during} diffusion is essential. It keeps the model’s samples within the data manifold while progressively refining motion under task constraints.

\begin{table}[t]
\footnotesize
\begin{center}

\setlength{\tabcolsep}{2pt}
\resizebox{\linewidth}{!}{
\begin{tabular}{l c c c c c c c c c c c} 
\toprule
& \multicolumn{3}{c}{Hand} & \multicolumn{3}{c}{Object} & \multicolumn{3}{c}{Interaction} \\
\cmidrule(r){2-4} \cmidrule(r){5-7} \cmidrule(r){8-10}
  & W- $\downarrow$ & WA- $\downarrow$ & ACC $\downarrow$ & ADD $\uparrow$ & ADD-S $\uparrow$ & ACC $\downarrow$ & ADD $\uparrow$ & ADD-S $\uparrow$ & ACC $\downarrow$\\
% GT contact label &  {6.61} & 10.79 & 3.28 &  {0.55} &  {53.7} &  {72.2} &  {72.2} & 0.07 &  {56.1} &  {74.5} &  {74.5} & 0.07 \\
% w/o Guidance  & 9.34 & 15.92 & 4.59 & 0.76 & 44.3 & 61.0 & 61.0 &  {0.13} & 45.0 & 62.1 & 62.1 &  {0.13} \\
% w/o $\mathcal{L}_{\text{interaction}}$  & 6.70 & 10.47 &  {3.12} & 0.61 & 28.5 & 42.9 & 42.9 & 0.09 & 30.6 & 45.5 & 45.5 & 0.09 \\
% \ours (ours) & 6.67 &  {10.41} & 3.26 & 0.58 & 51.1 & 69.9 & 69.9 & 0.11 & 53.5 & 72.3 & 72.3 & 0.11 \\
\midrule
GT contact  & 10.79 & 0.55 & 6.61 & 53.7 & 72.2 & 0.07 & 56.1 & 74.5 & 0.07 \\
Gen+Opt  & 15.92 & 0.76 & 9.34 & 44.3 & 61.0 & 0.13 & 45.0 & 62.1 & 0.13 \\
 w/o $\mathcal{L}_{\text{inter}}$  & 10.47 & 0.61 & 6.70 & 28.5 & 42.9 & 0.09 & 30.6 & 45.5 & 0.09 \\
% \ours w/ DROID-SLAM  & \yy{low prio} \\
\ours   & 10.41 & 0.58 & 6.67 & 51.1 & 69.9 & 0.11 & 53.5 & 72.3 & 0.11 \\
\bottomrule
\end{tabular}
}
\vspace{-1em}
\caption{\textbf{Ablation Study.} We compare our full model on hand, object, and relative motion quality against variants that use ground-truth contact, not alternate diffusion and guidance step, or exclude the interaction objective. }
\vspace{-1em}
\label{tab:ablation}
\vspace{-1.5em}

\end{center}
\end{table}

\begin{table}[t]
\footnotesize
\begin{center}

\setlength{\tabcolsep}{2pt}
\resizebox{\linewidth}{!}{
\begin{tabular}{l l c c c c c c} 
\toprule
 & \quad Test Data & \multicolumn{3}{c}{HOT3D} & \multicolumn{3}{c}{H2O} \\
\cmidrule(r){3-5} \cmidrule(r){6-8} 
Train Data &  & ADD $\uparrow$ & ADD-S $\uparrow$ & ACC $\downarrow$  & ADD $\uparrow$ & ADD-S $\uparrow$ & ACC $\downarrow$  \\
\midrule
HOT3D & \ours (Ours) & 51.1 & 69.9& 0.11 & \gray{44.7} &	\gray{64.9} & \gray{0.23} \\
H2O & H2OTR \cite{cho2023transformer} & \gray{3.2}	&\gray{7.7} & \gray{25.4} & 62.1	& 77.1 & 0.48 \\
\bottomrule
\end{tabular}
}
\vspace{-1em}
\caption{\textbf{Zero-Shot Generalization.} Gray lines denote zero-shot test dataset while unshaded line denote in-domain test split.}
\vspace{-1em}
\label{tab:h2o}
\vspace{-1.5em}
\end{center}
\end{table}

\paragraph{How much does the interaction loss help?}
In Table~\ref{tab:ablation} Line 3, we show that the interaction objectives that capture the spatial relationships and relative dynamics conditioned on predicted contact is also important for producing faithful object motion and realistic hand-object interactions.

\paragraph{How well does the model generalize?}
In Table~\ref{tab:h2o}, we evaluate \ours zero-shot on the unseen H2O dataset~\cite{kwon2021h2o}. While \ours experiences a moderate performance drop, RGB-conditioned baselines like H2OTR~\cite{cho2023transformer} collapse catastrophically out-of-distribution. We attribute this robustness to our motion-space prior, which exhibits a smaller domain gap than the appearance-based representations used in prior work. While encouraging, scaling to more diverse training data and broader cross-dataset evaluation remain important directions for achieving truly generalizable reconstruction.

\subsection{Application: Hand-Guided HOI Planner}
Our framework is flexible beyond reconstruction: given a coarse hand trajectory $\bar{\bm{H}}$, picking and placing times (contact label $\bm{C}$), along with the object template, we can directly synthesize diverse hand-object interaction motions without any video input.
This could potentially enable a robot planner to enumerate candidate object trajectories for a specific coarse hand motion at a specific picking and placing timing.
Specifically, the coarse hand trajectory serves as the diffusion conditioning, while contact labels are injected at each guidance step via an L2 loss on binary contact predictions alongside $g_{\text{inter}}$ and $g_{\text{temp}}$. Please refer to our project page for video results.

\section{Discussion}

\paragraph{Limitations and Future Work.}
Our current framework reconstructs each hand-object pair independently. A natural next step is extending it to scene-level, multi-object reconstruction through joint diffusion and scene-level objectives.
The method also assumes a known object template, which could be relaxed by incorporating LLM-based retrieval~\cite{wu2024reconstructing} or template-free generation~\cite{ye2024g,fan2024hold}.
Finally, since our generative prior is trained on a single dataset, scaling it with recent large-scale hand-object datasets~\cite{lu2025humoto,kim2025parahome} would improve generalization and robustness.

\paragraph{Conclusion.}
We introduced \ours, a unified framework for reconstructing hand articulation and object trajectories from metric-SLAMed egocentric videos.
By combining a learned generative motion prior with visual and contact guidance, it achieves coherent and plausible reconstructions under challenging conditions.
We believe this framework offers a promising step toward scalable, scene-level hand-object understanding and generative modeling of human interactions in everyday environments.

{
    \small
    \bibliographystyle{ieeenat_fullname}
    \bibliography{main}
}

\clearpage
\setcounter{table}{4}
\setcounter{figure}{5}
\maketitlesupplementary
\appendix

In supplementary material, we provide further details on implementing network, full VLM prompt, and evaluation metrics (Sec.~\ref{sec:detail}). We also visualize more comparisons and results in supplementary videos. 

\section{Implementation Details}
\label{sec:detail}
\paragraph{Network Architecture.} We use a 4-layer transformer decoder with 4 attention heads for hand-to-object diffusion. The network is non-autoregressive, processing sequences jointly following~\cite{tevet2022human}, with 12.35M parameters. The diffusion variable $x$ is a 73-dimensional vector comprising: a 9D object state, a 2D bimanual contact indicator, and bimanual hand representations ($2 \times 31$D)---global orientation (3), translation (3), pose PCA (15), and shape parameters (10). All inputs are projected to a consistent latent dimension of 512.
The network is trained for 1{,}000{,}000 iterations using AdamW at a learning rate of $2\times10^{-4}$.
To reduce overfitting, we augment the object template by sampling a random canonical pose applying a random rotation and a small translation jitter---for each training window.

\paragraph{Training Loss.} 
In addition to the DDPM loss, we introduce auxiliary objectives to enhance realism. 
(1) \textit{Interaction Loss} ($\mathcal{L}_{\text{inter}}$) encourages realistic contact between the predicted hand-object motions and contact labels. 
It penalizes hand-object distances when contact is predicted and enforces near-rigid transport of contact points across consecutive contact frames~\cite{li2023object,li2024controllable}.  
Specifically, for each hand joint, we find its nearest object point $p^i$, rotate it by the object’s relative motion, and penalize deviation from its counterpart $p^{i+1}$,
\ie $\|\bm{R}^{i+1} (\bm{R}^{i})^{T} p^{i} - p^{i+1}\|$, where $\bm{R}$ is the object rotation.
(2) \textit{Consistency Loss} ($\mathcal{L}_{\text{const}}$) promotes agreement among hand features before and after MANO forward kinematics, 
$\|\bm{J}_\psi - \text{MANO}(\bm{\Gamma}_\psi, \bm{\Lambda}_\psi, \bm{\Theta}_\psi)\|_2$.  
(3) \textit{Temporal Smoothness} ($\mathcal{L}_{\text{smooth}}$) further penalizes large accelerations.

\paragraph{Running Time.} On a single NVIDIA RTX 6000 Blackwell GPU, our model processes a 150-frame clip in an average of 59.34 seconds. The inference time is dominated by the guidance step (59.06s), with the diffusion step requiring only 0.28s. This represents an orders-of-magnitude speedup over prior works such as~\cite{fan2024hold} (30 hours) and~\cite{ye2023ghop} (1 hour). The peak memory footprint is 14GB. VLM queries take 18.6s on average per image with GPT-5.

\paragraph{VLM Prompt.}
We prompt a VLM to label contact info, with additnoal in-context-learning exmaples. Full prompts are illustrated in Table in this appendix.

\paragraph{Evaluation Metrics.}
All metrics are computed on 150-frame clips, which correspond to the original sequence length in HOT3D-CLIP, in contrast to prior work~\cite{zhang2025hawor}, which typically evaluates on shorter 60–100 frame segments taken from the middle of the videos.

To compute W/WA-MPJPE, we align the predicted trajectory to the ground truth using an affine transformation (scale, rotation, translation) estimated from selected keypoints.
WA-MPJPE uses all joints from all frames, while W-MPJPE uses only the joints from the first two frames. Although trajectory error could be computed without alignment given ground-truth cameras, we follow the standard alignment protocol used in prior work~\cite{wham,slahmr,zhang2025hawor}.

For ADD/ADD-S of objects, we align predictions using the ground-truth camera poses.
For HOI ADD/ADD-S, we first globally align the hand trajectory (as in WA-MPJPE) and then evaluate object error in this aligned space.
The usual AUC ~\cite{hodavn2020bop,foundationposewen2024} threshold of 0.1 is overly strict for egocentric HOI due to severe occlusion, truncation, and out-of-view frames, leading to saturated low scores.
We therefore use a more permissive threshold of 0.3 to obtain a more informative evaluation.

\onecolumn

\begin{tcolorbox}[
    colback=white,
    colframe=black,
    title=\textbf{System Instruction},
    fonttitle=\bfseries\color{white},
    coltitle=white,
    sharp corners,
    boxrule=0.5pt,
    breakable % Allows the box to split across pages automatically
]
\label{tab:long1}
\begin{Verbatim}[fontsize=\small]
You are a precise visual classifier for hand-object contact detection in 
cluttered scenes.

CRITICAL CONSTRAINTS:
1. Each hand (left/right) can be in contact with AT MOST ONE object at a time.
2. "In contact" means direct physical touch: grasping, holding, pressing, or 
   any visible contact.
3. If a hand is not clearly touching any object, you must mark all objects as 0 
   for that hand.
\end{Verbatim}
\end{tcolorbox}

\begin{tcolorbox}[
    colback=white,
    colframe=black,
    title=\textbf{User Prompt Template},
    fonttitle=\bfseries\color{white},
    coltitle=white,
    sharp corners,
    boxrule=0.5pt,
    breakable
]
\begin{Verbatim}[fontsize=\small]
Analyze this image for hand–object contact (actual touching, not just reaching).
VISUAL GUIDANCE:
The image has been annotated with colored masks:
- GREEN dot = Left hand
- RED dot = Right hand
- Other COLORED masks = Candidate objects (each object has a unique color)
CANDIDATE OBJECTS (in order):
1. obj1
2. obj2
3. obj3
...
STRICT DEFINITION OF CONTACT:
For this task, contact means clear physical touching in this frame only.
Contact (label = 1) requires BOTH:
1. Mask intersection:
   - The hand mask and the object mask share some pixels or directly overlap 
     at the boundary (no visible gap).
2. Touching region:
   - The overlap is at a plausible touching area (finger tips, fingers, palm, 
     side of hand) on the visible surface of the object.
NO Contact (label = 0) in all of these cases:
- The hand is reaching toward, hovering above, or very close to an object with 
  a visible gap between masks.
- The hand is aligned in depth (e.g., above or behind the object) but the masks 
  do not intersect.
- The hand is in a pose that suggests future contact, but there is no current 
  touching in this single frame.
- There is only a tiny, ambiguous intersection (1–2 pixels) that could be noise 
  or occlusion. In such uncertain cases, choose 0 (no contact).
IMPORTANT:
- **Reaching or hovering is NOT contact.**
- **If you are unsure whether contact is happening, choose 0 (no contact).**
CONSTRAINTS (VALIDATION CHECK):
- Each hand can touch AT MOST ONE object.
  - Sum of left across all objects must be <= 1.
  - Sum of right across all objects must be <= 1.
- If a hand is not clearly touching any object, it should have 0 for all objects.
OUTPUT FORMAT:
Return only a JSON object in this exact format (no extra text):
{
  "obj1": {"left": 0, "right": 1},
  "obj2": {"left": 0, "right": 0},
  "obj3": {"left": 1, "right": 0}
}
Where:
- 1 = the specified hand is clearly touching that object in this frame.
- 0 = the specified hand is not touching that object in this frame.\end{Verbatim}
\end{tcolorbox}

\twocolumn

\end{document}